%% file: paper.tex
\documentclass[lettersize,journal]{IEEEtran}
\IEEEpubidadjcol
\usepackage{amsmath,amsfonts}
\usepackage{algorithmic}
\usepackage{algorithm}
\usepackage{array}
\usepackage[caption=false,font=normalsize,labelfont=sf,textfont=sf]{subfig}
\usepackage{textcomp}
\usepackage{stfloats}
\usepackage{url}
\usepackage{verbatim}
\usepackage{graphicx}
\usepackage{cite}
\usepackage{color}
\usepackage{bbm}
\usepackage{multirow}
\usepackage[table,xcdraw]{xcolor}
\usepackage{booktabs}
\usepackage{amssymb}
\usepackage{bbding}
\usepackage{makecell}
\usepackage{hyperref} 
\usepackage{tabularx}
\newcommand{\rf}[1]{{\textbf{\color{red}{#1}}}} 
\newcommand{\bd}[1]{{\color{blue}{\underline{#1}}}} 
\begin{document}

\title{SSP-IR: Semantic and Structure Priors for Diffusion-based Realistic Image Restoration}

\author{Yuhong Zhang, Hengsheng Zhang, Zhengxue Cheng, Rong Xie,~\IEEEmembership{Member,~IEEE}, Li Song,~\IEEEmembership{Senior Member,~IEEE}, Wenjun Zhang,~\IEEEmembership{Fellow Member,~IEEE}
\thanks{Yuhong Zhang, Hengsheng Zhang, Zhengxue Cheng, Rong Xie and Wenjun Zhang are with
Institute of Image Communication and Network Engineering, Shanghai Jiao Tong University, China (e-mail: rainbowow@sjtu.edu.cn; hs\_zhang@sjtu.edu.cn; zxcheng@sjtu.edu.cn; xierong@sjtu.edu.cn; zhangwenjun@sjtu.edu.cn).}
\thanks{Li Song is with Institute of Image Communication and Network Engineering, Shanghai Jiao Tong University and the MoE Key Lab of
Artificial Intelligence, AI Institute, Shanghai Jiao Tong University, China (email: song\_li@sjtu.edu.cn).}
}

\markboth{Journal of \LaTeX\ Class Files,~Vol.~14, No.~8, August~2021}%
{Shell \MakeLowercase{\textit{et al.}}: A Sample Article Using IEEEtran.cls for IEEE Journals}


\maketitle

\begin{abstract}
Realistic image restoration is a crucial task in computer vision, and diffusion-based models for image restoration have garnered significant attention due to their ability to produce realistic results. Restoration can be seen as a controllable generation conditioning on priors. However, due to the severity of image degradation, existing diffusion-based restoration methods cannot fully exploit priors from low-quality images and still have many challenges in perceptual quality, semantic fidelity, and structure accuracy. Based on the challenges, we introduce a novel image restoration method, SSP-IR. Our approach aims to fully exploit semantic and structure priors from low-quality images to guide the diffusion model in generating semantically faithful and structurally accurate natural restoration results. Specifically, we integrate the visual comprehension capabilities of Multimodal Large Language Models (explicit) and the visual representations of the original image (implicit) to acquire accurate semantic prior. To extract degradation-independent structure prior, we introduce a Processor with RGB and FFT constraints to extract structure prior from the low-quality images, guiding the diffusion model and preventing the generation of unreasonable artifacts. Lastly, we employ a multi-level attention mechanism to integrate the acquired semantic and structure priors. The qualitative and quantitative results demonstrate that our method outperforms other state-of-the-art methods overall on both synthetic and real-world datasets. Our project page is \href{https://zyhrainbow.github.io/projects/SSP-IR}{https://zyhrainbow.github.io/projects/SSP-IR}.

\end{abstract}


\section{Introduction}
\IEEEPARstart{I}{mage} restoration (IR) has always been a classic problem in digital image processing, aiming to convert low-quality (LQ) images into high-quality (HQ) images. Typical image restoration tasks include super-resolution (SR) \cite{dong2016accelerating,zhao2020efficient,chen2023dual}, deblurring \cite{kupyn2018deblurgan,kupyn2019deblurgan,chen2024hierarchical}, denoising \cite{zhang2017beyond,zhang2018ffdnet}, color enhancement \cite{zeng2020learning,zhang2023dual}, inpainting \cite{suvorov2022lama,lugmayr2022repaint,corneanu2024latentpaint}, and compression artifact removal\cite{fu2021jpeg, kawar2022jpeg}. Early research often focused on studying image restoration problems under the assumption of a known linear degradation type present in the image. However, real-world low-quality images usually contain complex and diverse degradation types, making these methods less effective in practical applications. Therefore, BSRGAN\cite{zhang2021bsrgan} and Real-ESRGAN\cite{wang2021realesrgan} have proposed new degradation synthesis methods to model the degradation process and achieve realistic image restoration (real-IR). As image restoration techniques have advanced, there is a significant increase in the expectation of the perceptual quality of restored images. The over-smoothed results produced by the abovementioned methods fail to meet these expectations. To address this problem, generative adversarial networks (GANs) have been employed to solve the real-IR problem \cite{liang2022dasr,chen2022femasr,liang2022ldl,xie2023desra,wei2024tcsvtrescale}. By using adversarial loss, the restoration models can generate perceptually realistic results. However, GAN-based real-IR methods often introduce unnatural visual artifacts \cite{liang2022ldl,xie2023desra}.

\begin{figure}[t]
\begin{center}
\includegraphics[width=\linewidth]{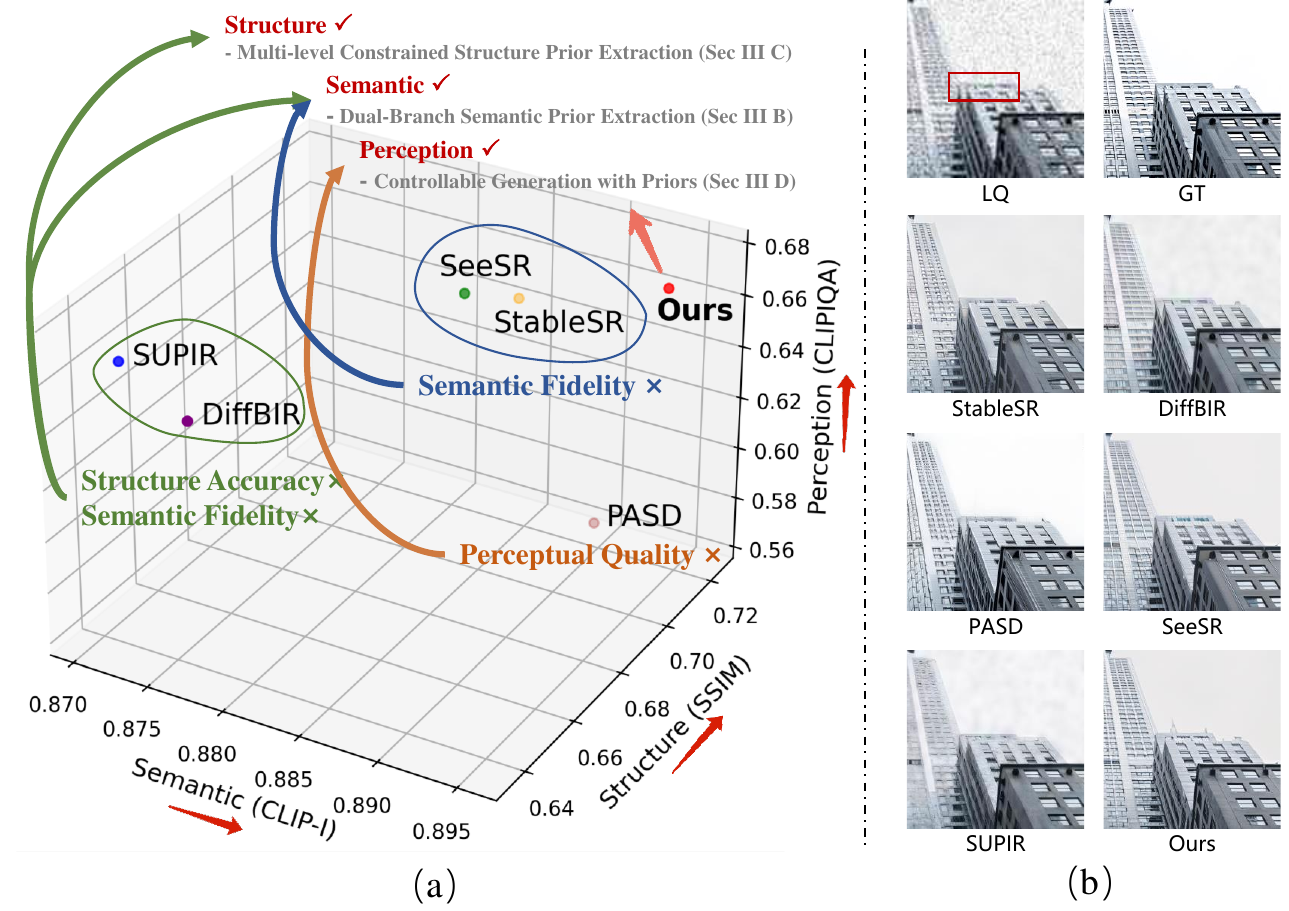}
\end{center}
\caption{Illustration of our method's superiority compared with other diffusion-based methods. (a) shows our method's good performance on the semantic, structure and perception quality. We calculate the CLIP similarity between the generated results and GT images to evaluate the semantic fidelity, SSIM to evaluate the structure accuracy and CLIPIQA to evaluate the perception quality. All metrics are calculated on RealSR dataset~\cite{cai2019realsr} and are the higher the better. Our method achieves a good balance about semantic-structure-perception. In the figure,  $\checkmark$ indicates better performance and $\times$ indicates poorer performance. (b) shows the visual comparison compared with other diffusion-based methods.}
\label{teaser}
\end{figure}

Recently, denoising diffusion probabilistic models (DDPMs) \cite{ho2020ddpm} have achieved remarkable success in the field of image generation, surpassing traditional generative adversarial networks (GANs) in various downstream tasks \cite{dhariwal2021diffusion}. As a result, researchers have started exploring the application of DDPMs for image restoration, incorporating high-quality generative priors from diffusion models into image restoration tasks\cite{kawar2022denoising,wang2022zero,saharia2022sr3,fei2023gdp,chen2024tcsvtface}. This approach aims to generate more realistic images without unnatural artifacts. Large-scale pre-trained text-to-image (T2I) models have gained great success. Particularly, stable diffusion (SD) which was trained on datasets containing over 5 billion image-text pairs, provides rich natural image priors and serves as a vast library of textures and structure. Some methods \cite{wang2023exploiting, lin2023diffbir,yang2023pixel,wu2023seesr,sun2023coser} have attempted to apply pre-trained stable diffusion to the blind image super-resolution task. 

Unlike traditional image restoration approaches, diffusion-based methods adhere to the principle of ``reconstruction as conditional generation", aiming to introduce control conditions to guide the diffusion model. DiffBIR~\cite{lin2023diffbir} trains a preprocessing model and introduces structure control to the diffusion model. However, this method requires two-stage training and lacks semantic prompts to assist the generation process. In addition, the preprocessing model model may lead to over-smoothing and reduce the structure accuracy. PASD~\cite{yang2023pixel} utilizes existing BLIP~\cite{li2022blip} to generate semantic prompts, yet the presence of artifacts in degraded images can lead to inaccurate extraction of semantic information as shown in Fig \ref{motivation}. SeeSR \cite{wu2023seesr} trains a dedicated RAM model \cite{zhang2023ram} to generate semantic tags, while CoSeR \cite{sun2023coser} trains a specialized cognitive encoder to generate object embedding. However, both SeeSR and CoSeR require separate training, and their performance depends on the additional training of RAM and cognitive encoders. They perform better when handling images with explicitly categorized objects but struggle with images lacking clear category information. 

With the advancement of multimodal technologies, Large Language Models (LLMs) or Multimodal Large Language Models (MLLMs) have played crucial roles in multimodal high-quality image generation and editing tasks \cite{koh2024lgie,jin2024llmra,fu2023mgie,huang2023smartedit}. SUPIR \cite{yu2024scaling} pioneers the application of text guidance from MLLM in restoration models, attempting to integrate MLLM and diffusion for image restoration. However, this method relies on prompts from pre-restored images, requiring additional training to generate these pre-restored images. Furthermore, the prompts are based on pre-restored images, which may not be faithful to the input. Moreover, SUPIR demands substantial training resources. Thus, the challenge of effectively leveraging prior conditions in degraded images to generate semantically faithful, structurally accurate, and perceptually high-quality images is a critical consideration.

In this paper, we present an effective solution \textbf{SSP-IR}, which fully exploits the semantic and structure priors in the input image for conditional generation. Specifically, for \textbf{Semantic Prior}, we adopt an explicit-implicit collaboration strategy to obtain richer and more accurate semantic information. We first utilize a multimodal large language model to generate detailed textual descriptions, which are explicit and can be manually adjusted by users to tailor the restoration results to meet requirements. Additionally, to acquire semantic information that is more faithful to the original image, we extract the image embedding of the input image with CLIP image encoder. We also design a Refine Layer to mitigate the influence of degraded artifacts. The explicit and implicit semantic information interact with each other to achieve a balance between fidelity and realism. Regarding \textbf{Structure Prior}, we introduce a constrained Processor to obtain degradation-independent structure prior. Finally, we integrate semantic and structure priors through \textbf{Prior-Guided Attention Module} to the denoising U-Net, achieving high-quality restoration results. Fig \ref{teaser} shows the superiority of our method in the quality of semantic, structure, and perception.

Our contributions can be summarized as follows:
\begin{itemize}
\item{We propose an effective one-stage realistic image restoration framework with semantic and structure priors to achieve high-quality restoration results.
\item{We introduce an explicit-implicit collaboration strategy to obtain accurate and specific semantic information by integrating the visual understanding capabilities of MLLM and the intrinsic representation capabilities of images.}}
\item{To ensure fidelity to input and avoid introducing degradation artifacts, we employ a preprocessing module with RGB and FFT constraints to extract degradation-independent structure features.}
\item{We employ a Prior-Guided Attention module to integrate the semantic and structure priors to the diffusion model. Qualitative and quantitative results on multiple datasets demonstrate the superiority of our method in terms of semantic, structure, and perception.}
\end{itemize}

\section{Related Work}
\subsection{Realistic Image Restoration}
The goal of realistic image restoration is to transform the LQ images with degradation artifacts to HQ and photo-realistic ones. Early research focused on image restoration of single degradation types, such as super-resolution \cite{dong2016accelerating,zhao2020efficient,chen2023dual,guan2024tcsvtfrequency, qiu2023tcsvtdual,chen2023tcsvtdynamic}, denoising \cite{zhang2017beyond,zhang2018ffdnet,ding2024tcsvtdenoisewavelet}, deblurring \cite{kupyn2018deblurgan, kupyn2019deblurgan,chen2024hierarchical}, and color enhancement \cite{zeng2020learning,zhang2023dual}. Some researchers are dedicated to studying baseline models for image restoration to handle various types of degradation\cite{liang2021swinir,chen2022nafnet,zamir2022restormer}, but training on specific datasets is required for each degradation type. Overall, these methods often rely on specific degradation types, while real-world degradation types are typically complex and diverse, making it challenging to apply these methods in practical scenarios. Therefore, some methods \cite{zhang2021bsrgan,wang2021realesrgan,dong2023panbsr} have made progress in the field of realistic image restoration by modeling the degradation process to simulate real-world image degradation. 

With the development of Generative Adversarial Networks, some studies\cite{zhang2021bsrgan,wang2021realesrgan,liang2022dasr,chen2022femasr,liang2022ldl,xie2023desra} have utilized pre-trained GANs to improve the super-resolution process, reducing the smoothness of generated images and enhancing the richness of texture details. Specifically, BSRGAN \cite{zhang2021bsrgan} utilizes a randomly selected degradation strategy to enhance the richness of degradation, while Real-ESRGAN \cite{wang2021realesrgan} employs a high-order degradation modeling process. DASR \cite{liang2022dasr} proposes an adaptive super-resolution network that estimates the degradation for each input image. FeMaSR \cite{chen2022femasr} utilizes a pre-trained VQGAN for feature matching between low-resolution features and the distortion-free high-resolution prior, resulting in more realistic and less artifact-prone image generation. However, GAN-based real-IR methods often introduce unnatural visual artifacts \cite{liang2022ldl,xie2023desra}. Therefore, in recent research, there is increasing interest in utilizing more advanced pre-trained generative models, such as denoising diffusion models \cite{ho2020ddpm,dhariwal2021diffusion, song2020denoising}, for image restoration to generate more realistic images.

\subsection{Diffusion-Based Image Restoration}
With the development of diffusion models, several attempts \cite{kawar2022denoising, wang2022zero, saharia2022sr3} have been made to utilize DDPM \cite{ho2020ddpm} for addressing image super-resolution problems. However, these methods often rely on simple linear degradation, making it challenging to apply them effectively in practical restoration scenarios. The assumption of known linear image degradation limits their practical application in complex scenes. 

Concurrently, other studies \cite{wang2023exploiting, lin2023diffbir, yang2023pixel, wu2023seesr, sun2023coser} have employed powerful pre-trained T2I models, such as stable diffusion \cite{rombach2022high}, to tackle real-world image restoration problems. Trained on billions of image-text pairs, stable diffusion possesses rich image priors that are crucial for real-world image restoration. StableSR \cite{wang2023exploiting} finetunes the SD model by training a time-aware encoder and an additional CFW module \cite{zhou2022codeformer} to balance fidelity and perceptual quality, enabling blind image restoration. DiffBIR \cite{lin2023diffbir} employs a two-stage strategy to address real-IR problems. It first reconstructs an image as an initial estimate and then employs SD priors to enhance image details. The aforementioned methods rely solely on images as conditions to activate the generation capability of T2I models. In contrast, PASD \cite{yang2023pixel} takes it a step further by utilizing high-level models, but the acquisition of high-level information relies on low-quality images. SeeSR \cite{wu2023seesr} trains a dedicated RAM model \cite{zhang2023ram} for assistance in restoration, while CoSeR \cite{sun2023coser} trains a dedicated cognitive encoder for object categorization. However, both SeeSR and CoSeR require additional training, and their restoration performance is dependent on the additional trained RAM and cognitive encoder.  Moreover, they perform better when dealing with images containing explicitly categorized objects, but struggle with images lacking clear category information. 

In conclusion, existing diffusion-based methods still have certain limitations in the real-IR task. Therefore, further exploration is needed on how to introduce semantic and structure information from low-quality images to more effectively leverage the potential of pre-trained T2I models in assisting real-IR.

\subsection{MLLM Asssited Image Restoration}
With the rapid development of text-to-image techniques, multimodal image generation has attracted increasing attention. LLMs and MLLMs have provided a range of solutions for high-quality multimodal image generation and editing tasks \cite{koh2024lgie,jin2024llmra,fu2023mgie,huang2023smartedit}. However, only a few works \cite{mu2023clarifygpt,jin2024llmra,wu2024towards} have extended LLMs or MLLMs to image restoration tasks. Clarity ChatGPT \cite{mu2023clarifygpt} combines ChatGPT with various pre-trained image restoration models, utilizing the rich prior reasoning capabilities of LLMs/MLLMs to determine, select, and invoke appropriate image restoration base models. LLMRA \cite{jin2024llmra} utilizes pre-trained MLLMs to generate detailed descriptions of the input degraded images. These descriptions are encoded as text features and serve as guiding conditions for transformer-based image restoration. \cite{wu2024towards} employs MLLMs to simultaneously process input images and chat-style user requests. It generates language dialogue responses as implicit guiding prompts, which are then used to guide image inpainting. However, these methods are only simple applications of MLLM and are not combined with the diffusion model to handle realistic image restoration. 

SUPIR \cite{yu2024scaling} applies text guidance from MLLMs in diffusion models. It combines the potential of MLLM and SD to realize image restoration, but the method relies on text descriptions generated on pre-restored images, which requires additional training to generate the pre-restored images. Moreover, the generated text descriptions are based on the pre-restored images, which may not be faithful to the inputs. In contrast, our approach directly utilizes the original image as input and leverages the visual understanding capability of MLLM to predict the content of the target restored image. This allows us to obtain text prompts that are more informative and faithful to the original image. In addition, we also utilize the image embedding of the input as a supplement, which is more conducive to generating the result faithful to the original image.

\section{Method}
\begin{figure*}[!htbp]
\begin{center}
\includegraphics[width=1.0\linewidth]{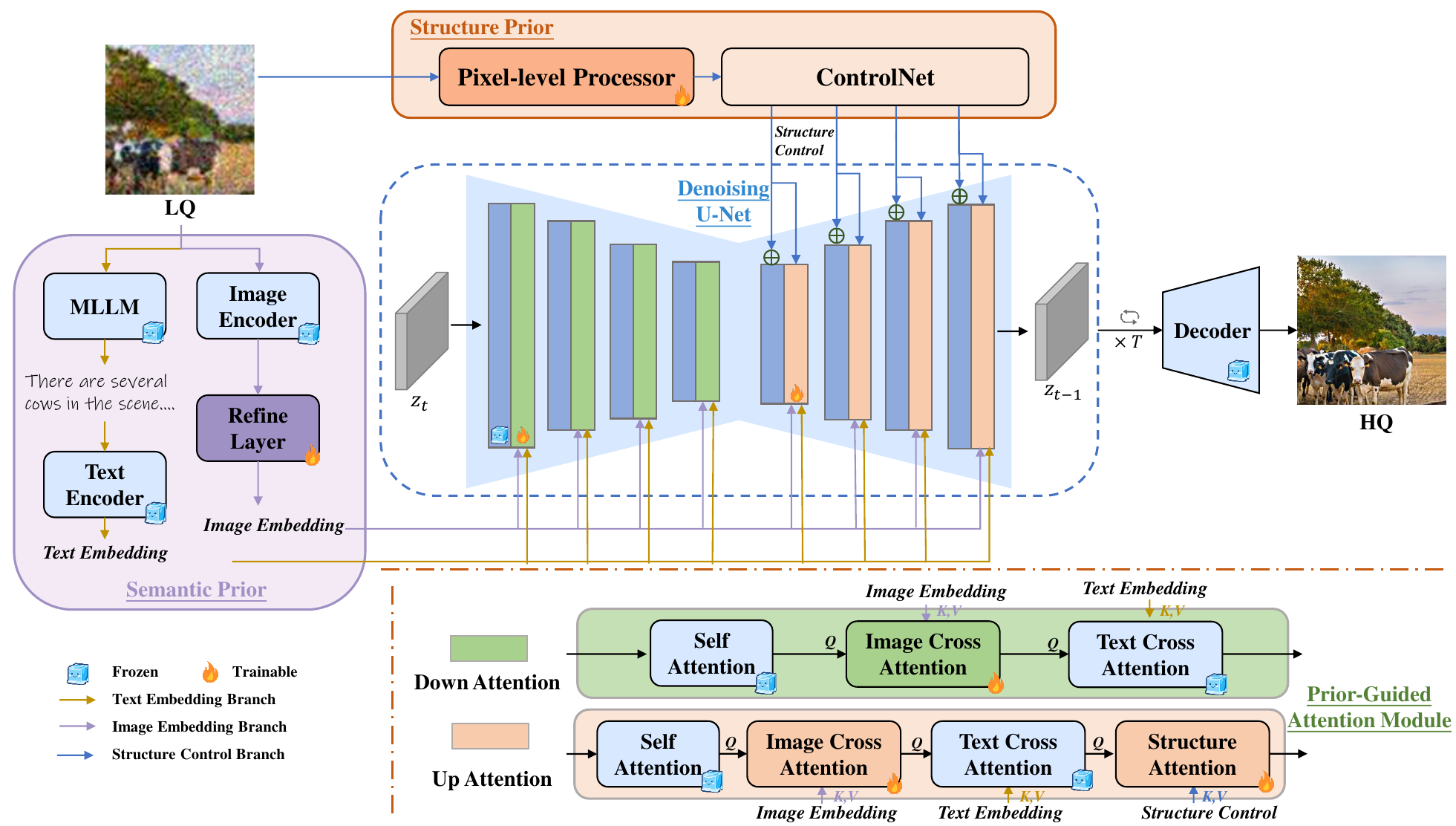}
\end{center}
\caption{The framework of the proposed method, which consists of semantic prior extraction module, structure prior extraction module and the denoising U-Net. Semantic prior extraction module consists of two branches and is used to generate explicit text embedding and implicit image embedding. Structure prior extraction module is used to generate degradation-independent structure control. The denoising U-Net integrates the text embedding, image embedding and strcuture control into the denoising process through the Prior-Guided Attention module.}
\label{framework}
\end{figure*}

In this section, we first describe the overall framework of our proposed SSP-IR. Then we describe how we get the semantic and structure priors and how we use the priors to the denoised U-Net in detail.

\subsection{Overview}
In this paper, we propose a diffusion-based realistic image restoration method with semantic and structure priors. We fully leverage the generative priors of pre-trained T2I models such as stable diffusion to generate realistic results. However, due to the uncontrollability of the diffusion process, we utilize semantic and structure information from low-quality images as guidance, aiming to generate realistic restoration results with faithful structures and clear details. The overall framework of our model is depicted in Fig \ref{framework}. We design specific modules to extract semantic and structure priors. The semantic and structure priors are integrated into the denoising U-Net through \textbf{Prior-Guided Attention modules}.

We design the \textbf{Semantic Prior} extraction module with two branches. \textbf{The first branch} utilizes the inference capability of a pre-trained multimodal language model to generate text embedding, which is explicit and can be modified by the user to suit their needs. \textbf{The second branch} utilizes the CLIP image encoder and a Refine Layer to generate the image embedding, which is implicit and contains the semantic information that cannot be expressed by text. The explicit text embedding and implicit image embedding collaborate as the semantic prior for the diffusion model's control.

To get the degradation-independent \textbf{Structure Prior}, we employ a constrained Processor and a ControlNet module. The Processor extracts degradation-independent features from the low-quality image under the supervision of RGB loss and FFT loss at different scales, while the ControlNet module utilizes these features to generate structure control in the latent domain.

Our model is trained end-to-end, unlike SeeSR~\cite{wu2023seesr} and CoSeR~\cite{sun2023coser}, which require additional training to obtain the semantic information from LQ images. With the aid of the powerful image understanding and reasoning capabilities of MLLM and CLIP, our method can directly extract meaningful semantic information from LQ images without the need for additional training.

\subsection{Explicit and Implicit Semantic Prior.}

\subsubsection{Analysis of the Influence of Semantic Prior}
\begin{figure}[!htbp]
\begin{center}
\includegraphics[width=0.9\linewidth]{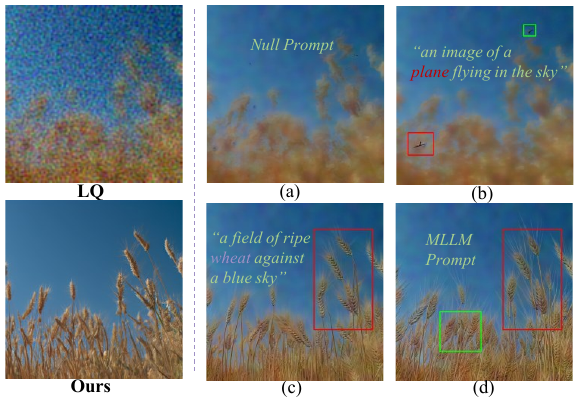}
\end{center}
\caption{The comparison of different prompts and their corresponding restoration results with PASD \cite{yang2023pixel}. (a) -(d) show the null prompt, BLIP prompt from LQ image, BLIP prompt from GT image and MLLM prompt predicted from LQ image and its corresponding restoration result. MLLM prompt is {\textit{``A field of ripe wheat stands against a clear blue sky, creating a picturesque sight. The golden wheat glows under the sunlight, while the serene blue sky provides a beautiful backdrop."}}}
\label{motivation}
\end{figure}

Semantic Prompts are key factors in determining the generation results of generative models. Therefore, how to mine semantic priors from low-quality images is an important issue we should consider. We use PASD \cite{yang2023pixel} as a baseline and employ null prompts, LQ prompts, and GT prompts to guide image restoration individually. The results are shown in Fig \ref{motivation}. We have two observations: 
\begin{itemize}
\item[(1)] Incorporating textual prompts contributes to generating details in images. The more details the prompts provide, the richer details will be generated. 
\item[(2)] Due to the influence of image degradation and the limitation of BLIP's caption generation capability, directly using the LQ prompts for image generation can lead to semantic errors (e.g., as shown in Fig \ref{motivation} (b), where an airplane is incorrectly generated).
\end{itemize}

Therefore, for realistic image restoration, we aim to infer reasonable and detailed semantic prompts from LQ images to guide the generation process. To get faithful and detailed semantic prompts, we propose a collaboration strategy to get explicit and implicit semantic prior from LQ images.

\subsubsection{Explicit-Implicit Strategy for Semantic Prior Extraction}
In concrete, benefiting from multimodal large language models, which possess extensive knowledge and powerful perception and reasoning capabilities, we utilize an MLLM to predict reasonable content prompts for potential blind image restoration results. Large language models can generate promising responses to user queries based on image information. We employ the pre-trained multi-modal large language model  LLaVA-7B \footnote{https://huggingface.co/liuhaotian/llava-v1.5-7b} to predict potential content representations from low-quality images, generating coherent and detailed textual descriptions and encoding them as text embeddings. The aforementioned process utilizes pre-trained models and does not require fine-tuning. The process can be summarized as follows:
\begin{equation}
\label{t_text}
T_{text} = E_{text}(E_{MLLM}(I_{LQ},T_{input}))
\end{equation}
where $T_{text}$ and $I_{LQ}$ denote the text embedding and the input low-quality image respectively, $T_{input}$ denotes the text instruction to the MLLM, such as \textit{``Describe the image in a very detailed manner if we remove the degradation artifacts from the image."} and $E_{text}$ and $E_{MLLM}$ indicates the CLIP text encoder and MLLM model respectively. MLLM can take both image and text sequences as input and generate coherent descriptions as output. Compared to predicting semantic tags \cite{wu2023seesr} and cognitive embeddings \cite{sun2023coser}, the predicted descriptions are more detailed. In addition, the semantic prior obtained in this way is explicit and conducive to human feedback.

The inference results of the MLLM include fine-grained descriptions. To ensure that the generated image remains semantically faithful to the original low-quality image, we introduce an image embedding branch to extract visual information as a supplement. Specifically, we use CLIP image encoder to extract additional image embedding. To eliminate information about degradation artifacts present in the image embedding, we design a Refine Layer to correct the image embedding, generating image embedding that is semantically faithful to the original low-quality image but free from degradation information. The process can be expressed as follows:
\begin{equation}
\label{t_image}
T_{image} = E_{refine}(E_{image}(I_{LQ}))
\end{equation}
where $T_{image}$, $E_{image}$, and $E_{refine}$ represent the image embedding, CLIP image encoder, and the Refine Layer respectively. The Refine Layer contains three MLP layers with Layer Normalization and LeakyReLU activations (except for the final layer). The semantic prior obtained in this way is implicit and can better reflect the semantic information in the input image that cannot be represented by explicit text.

Ultimately, the inferred explicit text embedding ($T_{text}$) and implicit image embedding ($T_{image}$) collaborate as the semantic prior to control the restoration process of the diffusion model.

\begin{figure*}[!htbp]
\begin{center}
\includegraphics[width=1.0\linewidth]{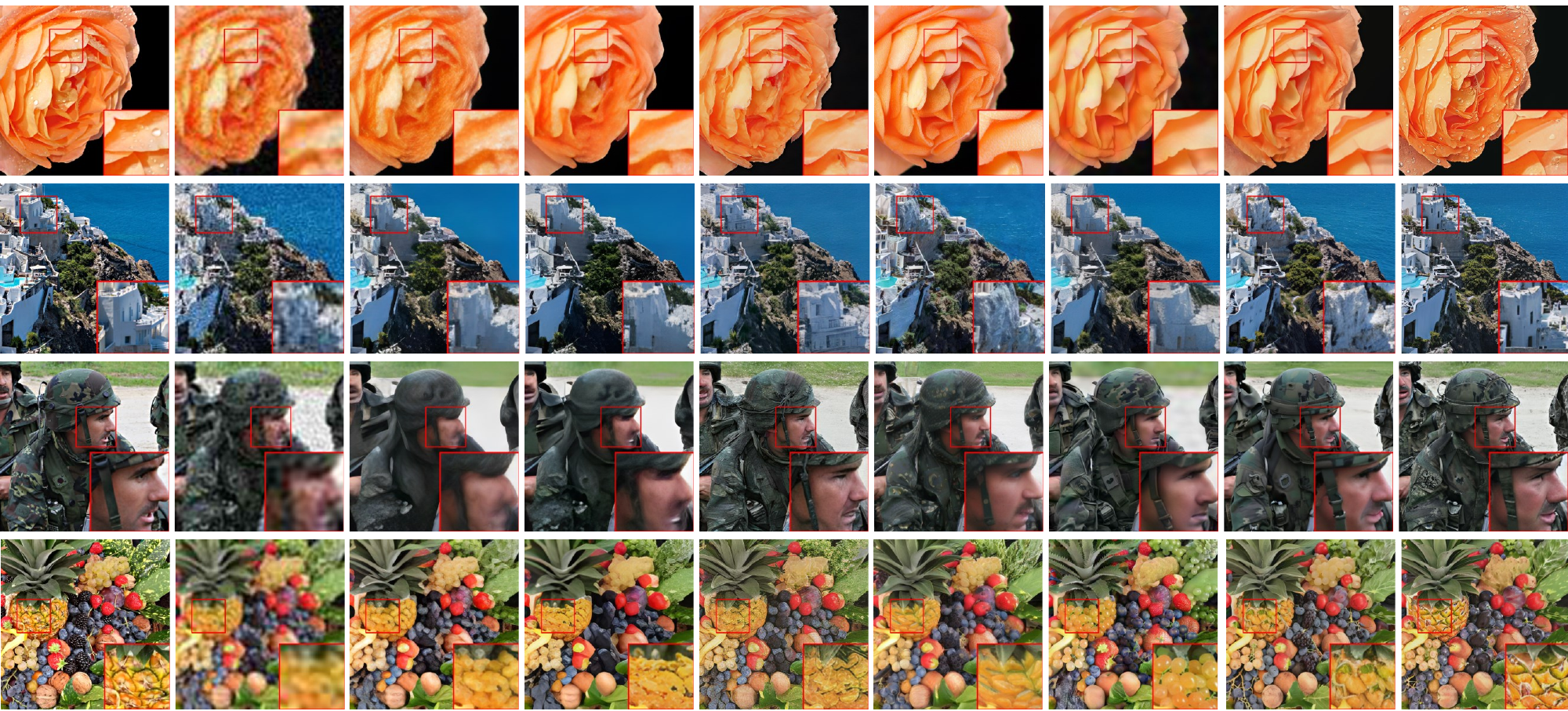}
{\begin{minipage}{\linewidth}
\centering
\vspace{-4mm}
\begin{tabularx}{\linewidth}{*{9}{>{\centering\arraybackslash}X}}
    \fontsize{7pt}{\baselineskip}\selectfont \textbf{GT} & \fontsize{7pt}{\baselineskip}\selectfont \textbf{LQ} & \fontsize{7pt}{\baselineskip}\selectfont \textbf{BSRGAN}\cite{zhang2021bsrgan} & \fontsize{5pt}{\baselineskip}\selectfont \textbf{Real-ESRGAN}
    \cite{wang2021realesrgan} & \fontsize{7pt}{\baselineskip}\selectfont \textbf{StableSR}\cite{wang2023exploiting} & \fontsize{7pt}{\baselineskip}\selectfont \textbf{DiffBIR}\cite{lin2023diffbir} & \fontsize{7pt}{\baselineskip}\selectfont \textbf{PASD}\cite{yang2023pixel} & \fontsize{7pt}{\baselineskip}\selectfont \textbf{SeeSR}\cite{wu2023seesr} & \fontsize{7pt}{\baselineskip}\selectfont \textbf{Ours}
\end{tabularx}
\end{minipage}}
\end{center}
\vspace{-6mm}
\caption{Qualitative comparisons with different state-of-the-art methods on synthetic datasets. We have provided the magnified regions for clarity. {\bf{Please zoom in for a better view.}}}
\label{compare_synth}
\end{figure*}

\subsection{Structure Prior Extraction with Constrained Processor}
Image restoration aims to recover high-quality authentic images from low-quality images. Different from general generation tasks, it requires the generated image to be structurally faithful to the low-quality image. However, due to real-world LQ images often suffer from complex and unknown degradation, directly extracting features from LQ images to control the diffusion process inevitably leads to the generation of images with undesirable artifacts. Therefore, we propose a multi-level constrained Processor to mitigate the impact of degradation and obtain degradation-independent latent structure features from the input LQ images for controlling the diffusion process. The Processor is a tiny network of four convolution layers to convert image-based conditions to 64 × 64 feature space to match the diffusion process. ControlNet \cite{zhang2023controlnet} utilizes these latent features to generate structure control signals in the latent domain, which are used to control the denoising process of the diffusion model. The process can be formulated as follows:
\begin{align}
F &= E_{processor}(I_{LQ})\label{eq:1} \\
\{P\}_{i=1}^4 &= E_{ControlNet}(F)\label{eq:2}
\end{align}
where $I_{LQ}$, $F$, and $P_i$ represent the LQ image, preprocessing feature and structure control respectively, and $E_{processor}$, $E_{ControlNet}$ represent the Processor and ControlNet module respectively.

What's more, to ensure the effectiveness of the processing operation, we use the multi-level RGB loss $\mathcal{L}_{RGB}$ and FFT loss $\mathcal{L}_{FFT}$ for supervision. Specifically, we extract intermediate features with 1/2, 1/4, and 1/8 resolutions ($F_{1/2}, F_{1/4}, F_{1/8}$) from the processor. Similar to PASD \cite{yang2023pixel}, we employ a convolutional layer to transform the feature maps at each scale into three-channel RGB space and compute the L1 distance with the corresponding high-quality images. To make further supervision and narrow gaps in the frequency domain, we make the FFT transform and calculate the loss in the frequency domain. The loss functions are computed as follows:
\begin{align}
\mathcal{L}_{RGB} &= \sum_{i=1/2,1/4,1/8}||I_i-I_{i}^{gt}||_1 \label{eq:loss_rgb}\\
\mathcal{L}_{FFT} &= \sum_{i=1/2,1/4,1/8}||FFT(I_i)-FFT(I_{i}^{gt})||_1 \label{eq:loss_fft}
\end{align}
where $I_i$ denotes the RGB image transformed from $F_i$, $I_{i}^{gt}$ denotes the ground truth at scale $i$ and $FFT$ denotes the Fast Fourier transform. In summary, $\mathcal{L}_{RGB}$ and $\mathcal{L}_{FFT}$ implement constraints on different scales and dimensions of the processor, which helps to extract degradation-independent structure features.

\subsection{Prior-Guided Attention Module}
Based on the above semantic and structure priors extraction modules, we have obtained text embedding $T_{text}$, image embedding $T_{image}$, and structure control $\{P_i\}_{i=1}^4$. All the prior information will be used to guide the diffusion model's denoising process to generate the desired restoration results. 

We design a Prior-Guided Attention module to integrate the priors. The Prior-Guided Attention module is shown at the bottom of Fig \ref{framework} and has a little difference in the Down Attention and Up Attention. The Down Attention represented in green includes self-attention, image cross attention, and text cross attention while the Up Attention represented in orange includes self-attention, image cross attention, text cross attention, and structure attention. Following the setting of the ControlNet, we only introduce the structure attention in the Up Attention of the denoising U-Net. Additionally, different from the original ControlNet which only introduces structure control by addition, we introduce structure attention to integrate structure control into the denoising process. This helps the generated images to be more faithful to the input images~\cite{yang2023pixel}. Text embedding is integrated into the diffusion process through text cross attention, leveraging the pre-trained text-to-image generation model's prior knowledge. To enable collaboration between text embedding and image embedding, we introduce another image cross attention to integrate image embedding into the denoising U-Net. 

For the attention mechanism, we use $Q$, $K$, and $V$ to represent the query, key, and value features, respectively. For structure attention, $Q$ comes from the denoising U-Net's features, while $K$ and $V$ come from the structure control $P_i$. The $K$ and $V$ of text cross attention and image cross attention come from text embedding and image embedding, respectively. The Prior-Guided Attention module effectively integrates the semantic and structure priors to the diffusion model to generate HQ images.

\begin{figure*}[!htbp]
\begin{center}
\includegraphics[width=1.0\linewidth]{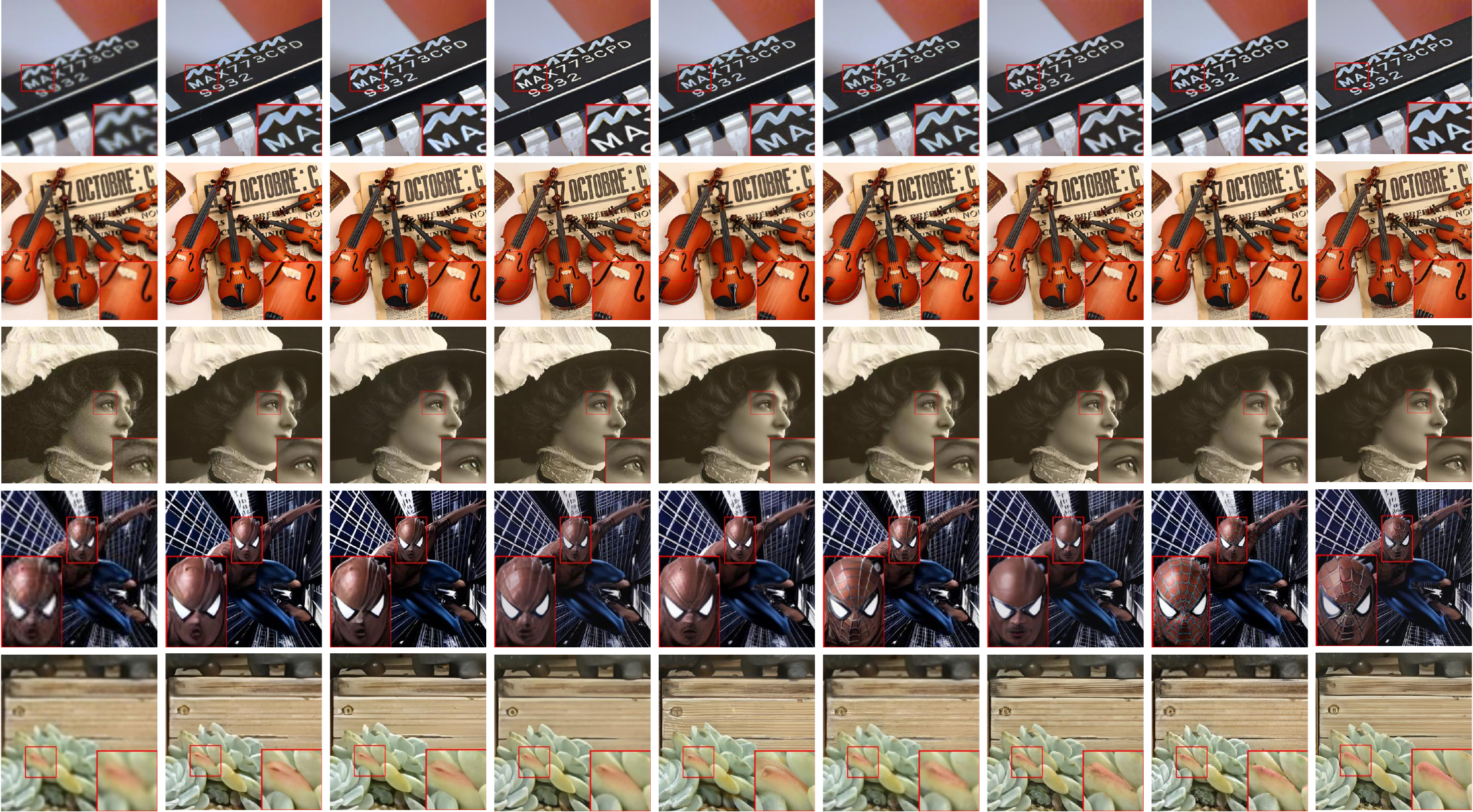}
{\begin{minipage}{\linewidth}
\centering
\vspace{-2mm}
\begin{tabularx}{\linewidth}{*{9}{>{\centering\arraybackslash}X}}
     \fontsize{7pt}{\baselineskip}\selectfont \textbf{LQ} & \fontsize{7pt}{\baselineskip}\selectfont \textbf{BSRGAN}\cite{zhang2021bsrgan} & \fontsize{5pt}{\baselineskip}\selectfont \textbf{Real-ESRGAN}\cite{wang2021realesrgan} & \fontsize{7pt}{\baselineskip}\selectfont \textbf{StableSR}\cite{wang2023exploiting} & \fontsize{7pt}{\baselineskip}\selectfont \textbf{DiffBIR}\cite{lin2023diffbir} & \fontsize{7pt}{\baselineskip}\selectfont \textbf{PASD}\cite{yang2023pixel} & \fontsize{7pt}{\baselineskip}\selectfont \textbf{SeeSR}\cite{wu2023seesr} & \fontsize{7pt}{\baselineskip}\selectfont \textbf{SUPIR}\cite{yu2024scaling} & \fontsize{7pt}{\baselineskip}\selectfont \textbf{Ours}
\end{tabularx}
\end{minipage}}
\end{center}
\vspace{-6mm}
\caption{Qualitative comparisons with different state-of-the-art methods on real-world datasets. We have provided the magnified regions for clarity. {\bf{Please zoom in for a better view.}}}
\label{compare_real}
\end{figure*}

\subsection{Training and Inference Strategies}
During the training process, the HQ image is encoded by the pre-trained VAE encoder \cite{rombach2022high} to obtain the latent code $z_0$. The diffusion process progressively adds noise to $z_0$ to generate $z_t$, where $t$ represents the random-sampled diffusion steps. By controlling the diffusion step $t$, the LQ image $I_{LQ}$, and the MLLM predicted prompt $T_{MLLM}$, we train the proposed diffusion-based restoration network, denoted as $\epsilon_{\theta}$, to estimate the noise added to the noise layer $z_t$. Our optimization objective of the diffusion process is:
\begin{equation}
\label{lossdiffusion}
\mathcal{L}_{diff} = \mathbbm{E}_{z_0,t,T_{MLLM},I_{LQ},\epsilon \sim \mathcal{N}}\left[||\epsilon-\epsilon_{\theta}(z_t,t,T_{MLLM},I_{LQ})||_2^2\right]
\end{equation}

Collaborating with \eqref{eq:loss_rgb} and \eqref{eq:loss_fft}, the final loss function is
\begin{equation}
\label{loss}
\mathcal{L} = \mathcal{L}_{diff} + \lambda_{1}\mathcal{L}_{RGB} + \lambda_{2}\mathcal{L}_{FFT}
\end{equation}
where $\lambda_{1}$ and $\lambda_{2}$ are balancing parameters. 

Specifically, the original CLIP text encoder in SD only encodes 75 tokens and cannot handle the entire generation prompts of MLLM. Therefore, we repeatedly use the text encoder to guide the encoding of all prompts. As illustrated in Fig.\ref{framework}, we freeze the parameters of the original SD model and only train the newly introduced modules, which include the Constrained Processor, ControlNet, and the newly added image attention and structure attention within the attention module to minimize training costs.

During the inference process, we employ the classifier-free guidance strategy \cite{ho2022classifier} and LRE strategy \cite{wu2023seesr}.

\section{Experiments}
\subsection{Experimental Settings}
{\bf{Implementation Details.}} We employ the pre-trained SD 1.5 \footnote{https://huggingface.co/runwayml/stable-diffusion-v1-5} model as the base pre-trained T2I model. During training, we finetune the model with Adam \cite{kingma2014adam} optimizer for 100K iterations. The batch size and the learning rate are set to 32 and 5e-5. For the hyperparameters in the loss function, we set $\lambda_{1}$ and $\lambda_{2}$ equal to 0.1 and 0.01 empirically. All experiments are conducted on an NVIDIA A100 GPU. For inference, we adopt DDPM sampling \cite{nichol2021improved} with 50 timesteps. 

{\bf{Training and Testing Datasets.}}
We train our model on DIV2K \cite{agustsson2017div2k}, Flickr2K \cite{timofte2017flickr2k}, OST \cite{wang2018ost}, the 9K face images from
FFHQ \cite{karras2019ffhq} and the 20K images from LSDIR \cite{li2023lsdir}. We use the degradation pipeline of Real-ESRGAN \cite{wang2021realesrgan} to synthesize
LQ-HQ training pairs. The pipeline consists of a second-order degradation process and each process randomly adds blur, resize, noise, and JPEG compression artifacts. The blur degradation involves isotropic Gaussian and anisotropic Gaussian kernels. The resize degradation involves area resize, bilinear interpolation and bicubic resize. The noise degradation involves additive Gaussian noise and Poisson noise. To conduct a comprehensive and reliable evaluation of the real-IR task, we conduct testing using synthetic and real-world datasets. For the synthetic test dataset, we create two datasets: DIV2K-Val and LSDIR-Val, which are generated from the DIV2K validation set \cite{agustsson2017div2k} and the LSDIR validation set \cite{li2023lsdir}, respectively.
For each synthetic dataset, We randomly crop 2K patches from the original validation set following the same degradation pipeline utilized during training. For the real-world test datasets, we employ the two widely used datasets, RealSR \cite{cai2019realsr} and DrealSR \cite{wei2020drealsr} for evaluation and use the same configuration as \cite{wang2023exploiting} to center-crop the LQ image to 128 × 128. What's more, We also use another real-world dataset, named RealLR200 \cite{wu2023seesr}, which comprises 200 LQ images that have been widely used in recent works and cover a diverse range of application scenarios. 

\input{tables/table1}

{\bf{Evaluation Metrics.}}
To provide a comprehensive and effective quantitative evaluation of the different methods, we employ a range of widely used reference and non-reference metrics, including PSNR, SSIM\cite{wang2004ssim} (calculated on the Y channel in YCbCr space), LPIPS\cite{zhang2018lpips}, FID\cite{heusel2017fid}, DISTS\cite{ding2020dists}, CLIP-I\cite{radford2021CLIP}, NIQE\cite{zhang2015niqe}, MANIQA\cite{yang2022maniqa}, MUSIQ\cite{ke2021musiq}, and CLIPIQA\cite{wang2023clipiqa}. For reference metrics, PSNR and SSIM are used to assess the similarity between generated images and ground truth images, while LPIPS and DISTS are employed to evaluate the perceptual quality of the generated images. CLIP-I estimates semantic accuracy by evaluating CLIP scores between generated images and ground truth images. For non-reference metrics, NIQE assesses the naturalness of images by utilizing statistical features derived from natural images. MANIQA evaluates the quality of images from both global and local perspectives using a multi-dimensional attention network. MUSIQ captures multi-scale image representations to provide a comprehensive assessment. CLIPIQA leverages the pre-trained vision-language model CLIP to evaluate image perceptual quality based on textual descriptions and visual features. These metrics offer objective measurements to evaluate the performance of different methods in terms of image similarity and perceptual quality, enabling a comprehensive assessment of the generated images. Non-reference metrics, in particular, are capable of better measuring the perceptual quality of images. We employ the IQA-Pytorch \footnote{https://github.com/chaofengc/IQA-PyTorch} to
compute these metrics.

\input{tables/table2}

{\bf{Compared Methods.}}
We compare our method with several state-of-the-art methods, which include GAN-based methods (BSRGAN \cite{zhang2021bsrgan}, Real-ESRGAN \cite{wang2021realesrgan}, DASR \cite{liang2022dasr} and FeMaSR) and diffusion-based methods (LDM \cite{rombach2022high}, StableSR \cite{wang2023exploiting}, DiffBIR \cite{lin2023diffbir}, PASD \cite{yang2023pixel} and SeeSR \cite{wu2023seesr}). Additionally, We also make a comparison with the MLLM assisted method SUPIR on the three real-world datasets. We use the released version of these competing methods for testing.

\subsection{Experimental Results}
{\bf{Qualitative Comparison.}}
 We first provide the qualitative comparison in Fig \ref{compare_synth} and Fig \ref{compare_real}. 
 For the synthetic datasets, as shown in Fig \ref{compare_synth}, GAN-based methods often produce smoother restoration results. When the image degradation is severe, they may fail to generate accurate results. In contrast, diffusion-based models can generate more detailed results. However, since StableSR and DiffBIR do not utilize semantic information to guide the diffusion model, their restoration results often lack richness in details and exhibit unclear textures. For example, the face structure (row 3) is unclear and the texture of the pineapple (row 4) is blurry. Although PASD and SeeSR take semantic information into consideration, they also produce incorrect restoration results due to inaccurate semantic predictions. As shown in Fig \ref{compare_synth}, PASD mistakenly identifies the pineapple as another fruit, resulting in an erroneous generated output (row 4) while SeeSR fails to correctly identify the building, resulting in the generation of rocks in place of the building (row 2). Additionally, benefiting from the image understanding capability of MLLM, our method is able to predict accurate semantic information from low-resolution images. Specifically, as shown in row 1, the restoration results of other methods do not include dewdrops on the flowers, while only our method's restoration results contain dewdrops. Moreover, the structure and texture of the flowers generated by our method are more defined and clearer. To be honest, due to significant information loss in degraded images, achieving image restoration results that are exactly the same as ground truth is challenging. The restored images obtained through our method only represent high-quality and plausible solutions.

 For the real-world datasets, as shown in Fig \ref{compare_real}, our method also generates results with reasonable structure, accurate semantics, and rich details such as the text on the chip (row 1), the string structure of the violin (row 2), the eye details (row 3), the texture details of Spider-Man's facial details (row 4) and the structure and clarity of the succulent (row 5). In contrast, SUPIR often generates excessive details, such as wrinkles in the eyes (row 3), surface of the succulent (row 5), which are not faithful to the input images.

In conclusion, our method demonstrates superior performances in terms of both the rationality of structures and the clarity of details, thanks to the powerful image understanding and reasoning capabilities of MLLM, as well as the application of image embedding and structure control.

\begin{figure}[tbp]
\begin{center}
\includegraphics[width=0.9\linewidth]{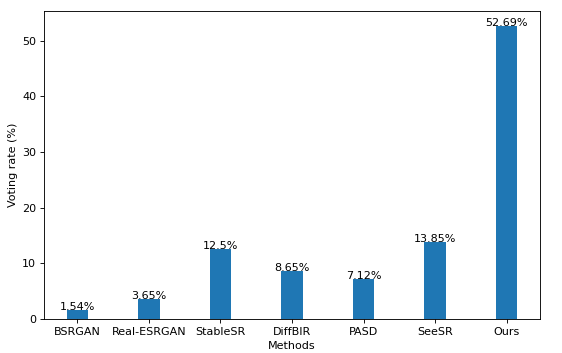}
\end{center}
\vspace{-4mm}
\caption{User study about voting rate of 7 methods on 26 images evaluated by 20 participants.}
\label{user_study}
\end{figure}

\input{tables/table3}

{\bf{Quantitative Comparison.}}
We then provide the quantitative comparison in Table \ref{tab:synthe} and Table \ref{tab:real}. Table \ref{tab:synthe} shows the quantitative results on the synthetic datasets and Table \ref{tab:real} shows the quantitative results on the real-world datasets. Compared with diffusion-based methods, our method achieves the best results in CLIP-I, MANIQA, MUSIQ, and CLIPIQA on all five datasets, except for the score of MUSIQ on RealLR200. This demonstrates the effectiveness of our proposed semantic prior extraction module and the superiority of our approach in terms of semantic accuracy and perceptual quality. What's more, our method outperforms other state-of-the-art methods in LPIPS, FID, and DISTS on the DIV2K-Val dataset and LSDIR-Val dataset. Compared with GAN-based methods, diffusion-based methods tend to getlower PSNR and SSIM scores. This is mainly because diffusion-based methods tend to generate more realistic details but at the expense of fidelity. While PSNR and SSIM are commonly used objective metrics for image quality assessment, they have limitations in capturing perceptual quality and fine details, as mentioned in \cite{yu2024scaling}. However, for the structural aspect, our method also obtains a relatively high SSIM score on all datasets compared with other diffusion-based methods, demonstrating the structure-preserving ability of the proposed method with the structure prior. Overall, our method achieves higher scores in no-reference metrics while maintaining competitive performance in reference metrics.

{\bf{User Study.}}
We conduct a user study with 20 participants to assess various methods through subjective evaluation. We compared our method with other six competitive restoration methods (BSRGAN, RealESRGAN, StableSR, DiffBIR, PASD, and SeeSR). We provided LQ images as references and asked the participants to select the best restoration result from the seven methods mentioned above. We randomly selected 26 images from synthetic and real-world datasets for testing and received evaluations from 20 participants, resulting in a total of 520 votes (26 images × 20 participants). As shown in Fig \ref{user_study}, our method significantly outperformed all other six competing methods, receiving 52.69\% of the votes, 38.84\% higher than the second place.

\begin{figure}[!tbp]
\begin{center}
\includegraphics[width=\linewidth]{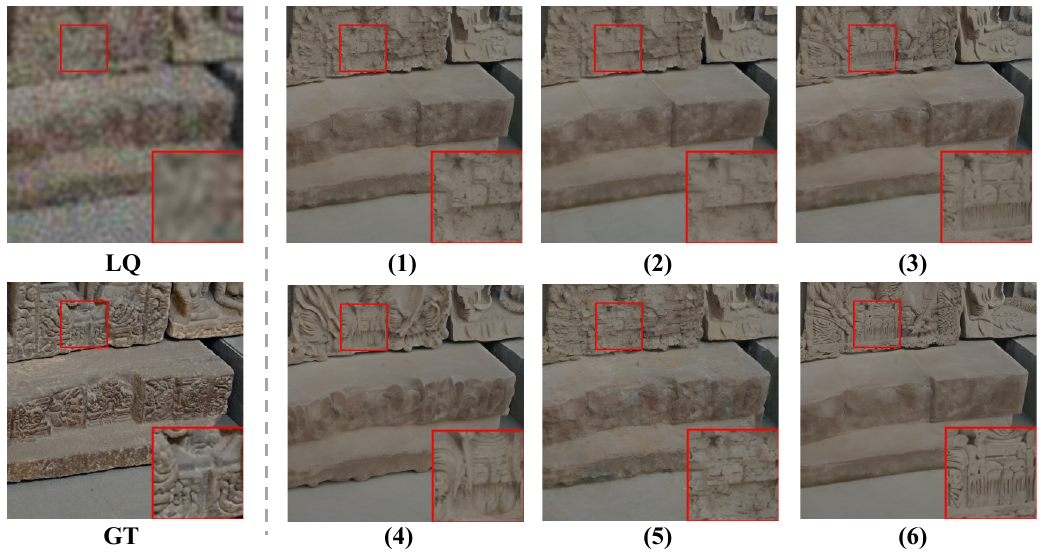}
\end{center}
\vspace{-4mm}
\caption{Visual comparison for the ablation study of different semantic prompts settings. (1) is the result with null prompt, (2) is the result with BLIP prompt, (3) is the result with null image embedding, (4) is the result of replacing the Refine Layer with only one linear layer, (5) is the result with null prompt and null image embedding, and (6) is the result of our method.}
\label{as_prompt}
\end{figure}

\subsection{Ablation Study}
To validate the effectiveness of the proposed method in detail, we conducted ablation experiments from the perspectives of {\bf{semantic prior}}, {\bf{structure prior}}, and {\bf{attention module}}. Additionly, we also validated the directional restoration capability of our proposed method driven by textual prompts.

Firstly, we analyze the influence of {\bf{semantic prior}}, specifically the impact of text embedding and image embedding. We conduct experiments considering the following five cases:
\begin{itemize}
\item[(1)]{ Replacing the prompt generated by MLLM with a null prompt.}
\item[(2)]{Replacing the prompt generated by MLLM with  the prompt generated by BLIP.}
\item[(3)]{Setting the image embedding to null.}
\item[(4)]{Replacing the Refine Layer module with only one linear layer to project the dimension of the CLIP-encoded image embedding to the image attention dimension.}
\item[(5)]{Setting both the image embedding and text embedding to null.}
\end{itemize}

\input{tables/table4}
\input{tables/table5}

\begin{figure*}[!htbp]
\begin{center}
\includegraphics[width=\linewidth]{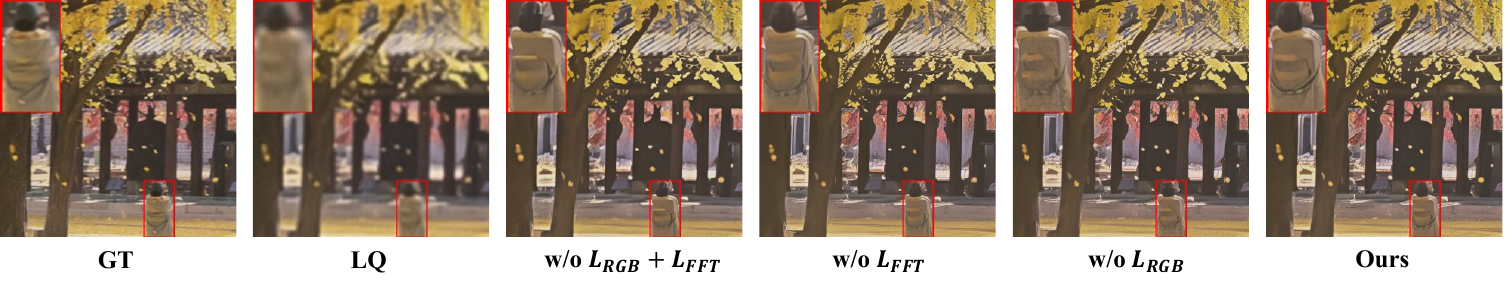}
\end{center}
\vspace{-4mm}
\caption{Visual comparison for the ablation study of different loss functions settings. With the constraints of RGB loss and FFT loss, the human's belt structure and clothing texture are more distinct.}
\label{as_loss}
\end{figure*}

\begin{figure*}[!htbp]
\begin{center}
\includegraphics[width=\linewidth]{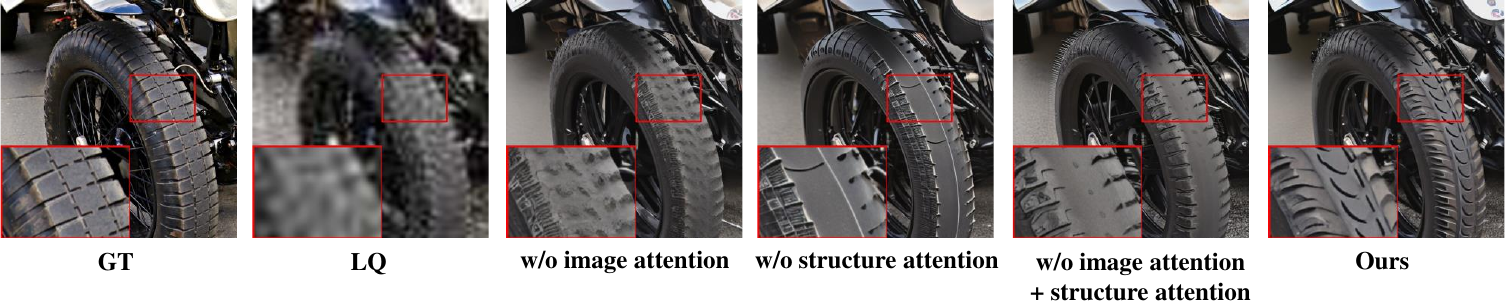}
\end{center}
\vspace{-4mm}
\caption{Visual comparison for the ablation study of different attention settings. With the image attention and structure attention, the tire tread pattern is more realistic and clearer.}
\label{as_attention}
\end{figure*}

\begin{figure}[!htbp]
\begin{center}
\includegraphics[width=\linewidth]{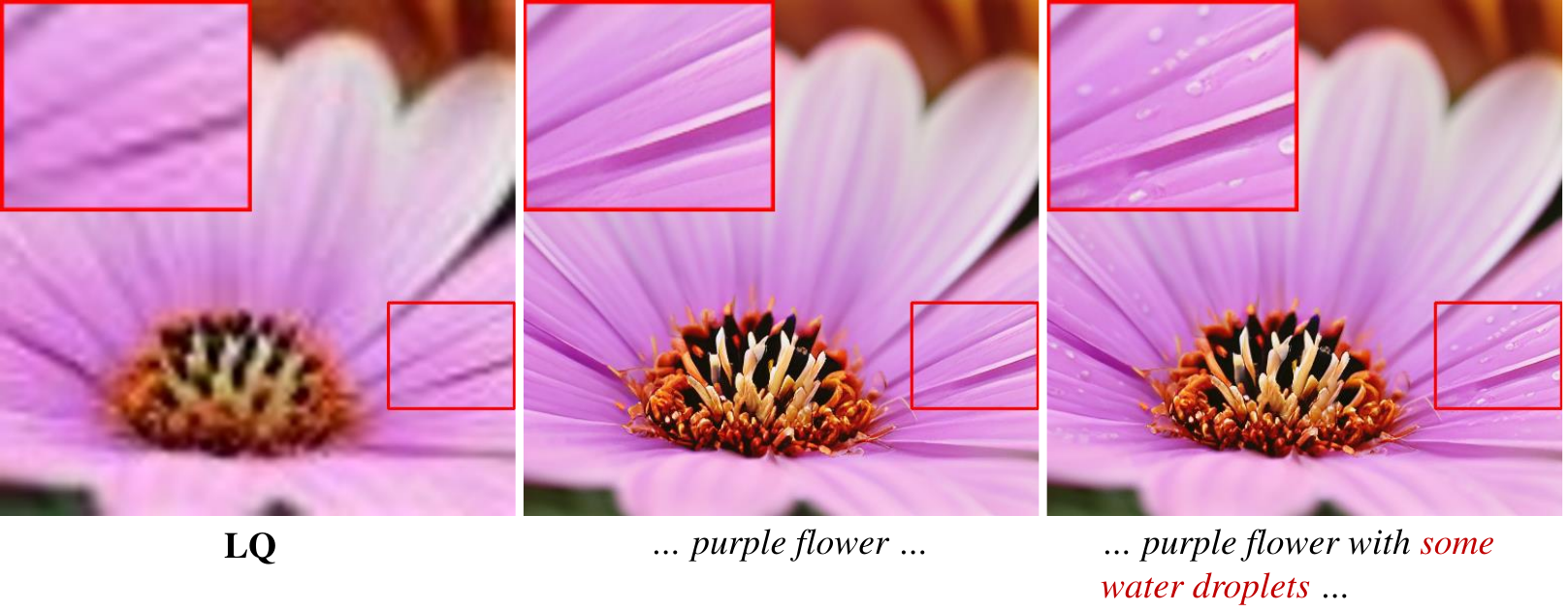}
\end{center}
\vspace{-3mm}
\caption{Visual comparison for directional restoration capability of different prompt. Users can modify the details of the restoration results as needed.}
\label{prompt_test2}
\end{figure}

The results are shown in Table \ref{tab:prompt} and Fig \ref{as_prompt}. Table \ref{tab:prompt} showcases the significant enhancement in metrics such as FID, LPIPS, CLIP-I, MUSIQ, and CLIPIQA by incorporating text embedding and image embedding. What's more, the PSNR metric of our method remains at a comparable level. Especially when setting both the image embedding and text embedding to null will lead to a significant decrease in CLIP-I, which fully proves the effectiveness of our proposed semantic prior. The quantitative metrics indicate that our approach excels in both semantic and texture generation, resulting in more realistic and accurate outputs. Additionally, it can be observed in Fig \ref{as_prompt} that setting both the text embedding and image embedding to null leads to the loss of texture details. Additionally, using BLIP-generated prompts as a replacement for MLLM-generated prompts may result in suboptimal restoration outcomes due to potential inaccuracies in the prompts generated from LQ images, such as the incorrectly generated airplane in Fig \ref{motivation} (b) and the engraved pattern on the stone in Fig \ref{as_prompt} (2).  Furthermore, when replacing the Refine Layer with only one linear layer, the lack of correction processes significantly decreases the restoration performance, specifically regarding the clarity of the statue texture highlighted by the red box in Fig \ref{as_prompt}, and leads to a decline in evaluation metrics as well.

\begin{figure*}[]
\begin{center}
\includegraphics[width=1.0\linewidth]{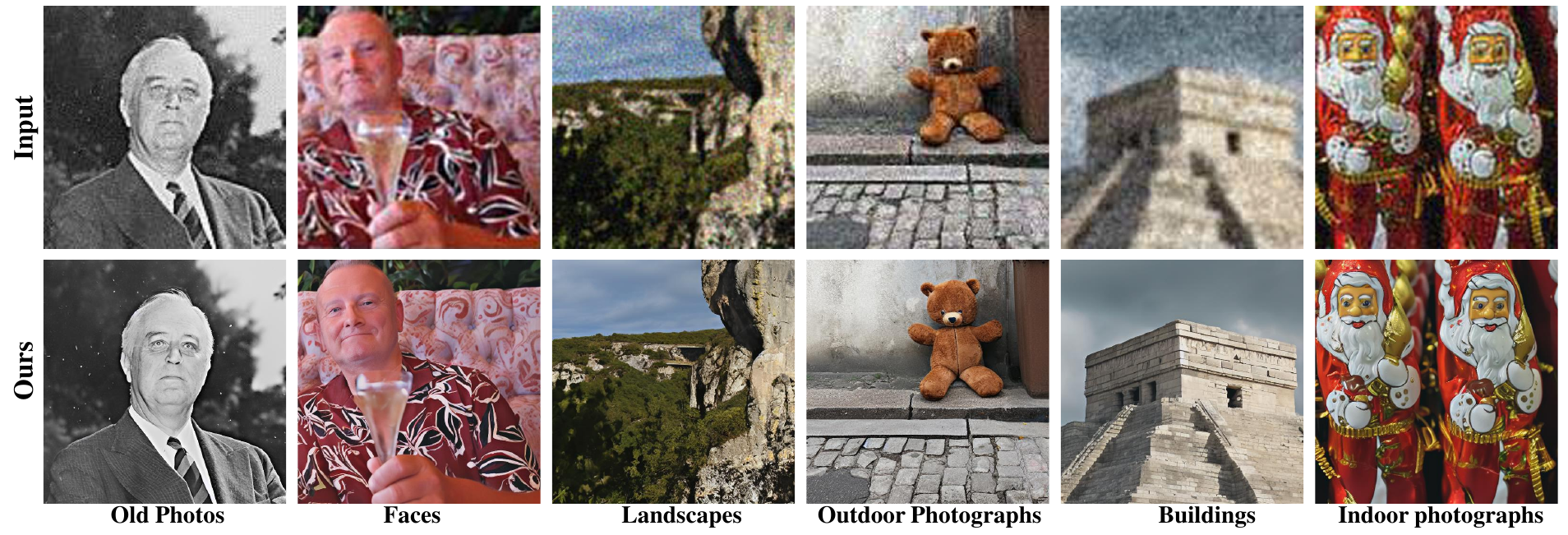}
\end{center}
\vspace{-2mm}
\caption{More examples to show the generalizability of the proposed method. Our method achieves great performance in various scenes.}
\label{general}
\end{figure*}

\input{tables/table_complexity}

Next, we evaluate the impact of loss constraints for {\bf{stucture prior}} extraction on the restoration results. The results are presented in Table \ref{tab:loss} and Fig \ref{as_loss}. It can be observed that removing $\mathcal{L}_{RGB}$ and $\mathcal{L}_{FFT}$ leads to an overall decline in the generation quality. Although removing $\mathcal{L}_{RGB}$ may yield gains in terms of MUSIQ and CLIPIQA metrics on the RealSR dataset, the generated details may contain artifacts and the structural details may not be sufficiently clear. As shown in Fig \ref{as_loss}, removing $\mathcal{L}_{RGB}$ and $\mathcal{L}_{FFT}$ results in artifacts on the person's hair, and the details of the clothing appear blurry. Removing $\mathcal{L}_{FFT}$ leads to an unclear structure of the belt on the person's clothing. Removing $\mathcal{L}_{RGB}$ causes overall blurry artifacts on the person's clothing. In contrast, our method performs better in terms of preserving the shape contours of the person and capturing clothing details.

Then, we evaluate the effectiveness of the introduced {\bf{image cross attention and structure attention}} in Prior-Guided Attention Module. The experimental results are shown in Table \ref{tab:attention} and Fig \ref{as_attention}. In terms of quantitative metrics, it is evident that the removal of image attention and structure attention leads to a significant decrease in PSNR, indicating a decline in the accuracy of the generated structural details. CLIP-I will be significantly reduced when image attention is removed. Simultaneously, the individual removal of image attention and structure attention impacts metrics such as FID and MUSIQ, which assess the quality of the generated images. Our method achieves a trade-off between realism and fidelity. Visually, it can be observed in Fig \ref{as_attention} that when removing image attention, the texture of the tire is not very clear. When removing structure attention, although the texture of the tire is clear, the structural details of the tire's texture are incomplete. When removing image attention and structure attention, both the texture structure and clarity are compromised. Our method achieves better results in terms of structure and clarity thanks to both image attention and structure attention being employed.

We validate the \textbf{directional restoration capability} of our proposed method. Fig \ref{prompt_test2} provides an example. Given an LQ image, the generated restoration result using our model is shown in the second column, which is consistent with the text prompt generated by MLLM. Since the text output by MLLM is explicit, users can further modify the text details for personalized restoration. As shown in the third column, by changing the text prompt, the details in the restoration image can be effectively modified, which enhances the application value of the model.

We also discuss the generalizability of the proposed method. We provide more visual examples to show our method's performance. As shown in Fig \ref{general}, our method shows great performance in various scenes, including old photos, faces, landscapes, outdoor and indoor photographs, and buildings. This proves that our method has wide application value.

Lastly, we analyze the complexity of the proposed method. The results are shown in Table \ref{tab:complexity}. Compared with other methods, our method only needs one-stage training and without any other additional training. Since we introduced a new attention module, we inevitably introduced more training parameters and longer inference time. But the trainable parameters are still in the scale of hundred million and affordable.

\section{Conclusion And Limitations}
In this paper, we propose a novel multimodal-perception approach with semantic and structure priors for realistic image restoration based on the diffusion model. Regarding semantic information, we consider both explicit and implicit perspectives. We utilize the multimodal large language model to infer reliable semantic text representations from low-quality images. Additionally, we use the CLIP image encoder and a designed Refine Layer to generate the image embeddings, which collaborate with the text embedding to achieve a balance between fidelity and realism. For structure control, we employ the multi-level constrained Processor and ControlNet to extract structure conditions. Lastly, we integrate the control information and the diffusion model using the Prior-Guided Attention Module. Our method demonstrates state-of-the-art performance on multiple datasets both qualitatively and quantitatively. 

To be honest, due to the utilization of the MLLM and diffusion model, our approach requires higher computational resources and longer inference time. However, We hope that this work serves as an inspiration for the application of MLLM in diffusion-based image restoration and provides valuable insights for future research endeavors. We acknowledge the need to explore more efficient and concise methods for image restoration and researching more efficient and concise methods for image restoration is also a direction we will pursue in the future.

\bibliographystyle{IEEEtran}
\bibliography{reference.bib}

\vfill

\end{document}

%% file: tables/table1.tex
\begin{table*}[!ht]
\caption{Quantitative comparison with different state-of-the-art methods on synthetic datasets. The best and second best results of each metric are highlighted in \rf{red} and \bd{blue}}
\label{tab:synthe}
\vspace{-3mm}
\resizebox{1.0\textwidth}{!}{
\begin{tabular}{c|c|cccc|cccccc}
\toprule
 &  & \multicolumn{4}{c|}{GAN-based Methods} & \multicolumn{6}{c}{Diffusion-based Methods} \\ 
\multirow{-2}{*}{Datasets} & \multirow{-2}{*}{Metrics} & BSRGAN\cite{zhang2021bsrgan} & Real-ESRGAN\cite{wang2021realesrgan} & DASR\cite{liang2022dasr} & FeMaSR\cite{chen2022femasr} & LDM\cite{rombach2022high} & StableSR\cite{wang2023exploiting} & DiffBIR\cite{lin2023diffbir} & PASD\cite{yang2023pixel} & SeeSR\cite{wu2023seesr} & Ours \\ \midrule
 & PSNR $\uparrow$ & \bd{20.04} & \rf{20.24} & 19.93 & 19.09 & 18.91 & 19.10 & 19.40 & 19.50 & 19.34 & 19.18 \\
 & SSIM $\uparrow$ & \bd{0.5134} & \rf{0.5352} & 0.5114 & 0.4781 & 0.4489 & 0.4414 & 0.4542 & 0.4935 & 0.4929 & 0.4918 \\
 & LPIPS $\downarrow$ & 0.4187 & 0.3913 & 0.4289 & 0.3991 & 0.4183 & 0.4121 & 0.4271 & 0.4827 & \bd{0.3883} & \rf{0.3840} \\
 & FID $\downarrow$ & 73.47 & 62.61 & 76.57 & 62.77 & 50.02 & 45.98 & 49.09 & 56.98 & \bd{39.94} & \rf{38.07} \\
 & DISTS $\downarrow$  & 0.2774 & 0.2642 & 0.2872 & 0.2459 & 0.2447 & 0.2581 & 0.2477 & 0.2725 & \bd{0.2284}  & \rf{0.2225} \\
 & CLIP-I $\uparrow$ & 0.7952 & 0.8340 & 0.7857 & 0.8125 & 0.8277 7 & 0.8544 & 0.8540 & 0.8094 & \bd{0.8698} & \rf{0.8982} \\
 & NIQE $\downarrow$ & 4.8133 & 5.0456 & 4.9213 & \rf{4.6122} & 5.8359 & \bd{4.7557} & 4.7727 & 5.5407 & 4.9767 & 5.0241 \\
 & MANIQA $\uparrow$ & 0.3513 & 0.3782 & 0.3156 & 0.3051 & 0.3481 & 0.3974 & 0.4525 & 0.4022 & \bd{0.5134} & \rf{0.5246} \\
 & MUSIQ $\uparrow$ & 58.78 & 58.48 & 54.22 & 57.74 & 59.31 & 62.60 & 64.59 & 59.15 & \bd{68.70} & \rf{69.27} \\
\multirow{-9}{*}{DIV2K-Val} & CLIPIQA $\uparrow$ & 0.5200 & 0.5532 & 0.5249 & 0.5669 & 0.5997 & 0.6535 & 0.6680 & 0.5561 & \bd{0.7062} & \rf{0.7091} \\ \midrule
 & PSNR $\uparrow$ & \rf{16.98} & \bd{16.96} & 16.86 & 16.25 & 16.49 & 16.28 & 16.73 & 16.62 & 16.55 & 16.30 \\
 & SSIM $\uparrow$ & 0.4122 & \rf{0.4325} & \bd{0.4114} & 0.3904 & 0.3826 & 0.3600 & 0.3845 & 0.3926 & 0.3965 & 0.3922 \\
 & LPIPS $\downarrow$ & 0.4533 & 0.4103 & 0.4626 & 0.4248 & 0.4403 & 0.4188 & 0.4227 & 0.4876 & \bd{0.3976} & \rf{0.3871} \\
 & FID $\downarrow$ & 69.00 & 56.59 & 70.74 & 60.41 & 46.11 & 39.58 & 38.72 & 49.72 & \bd{34.79} & \rf{31.85} \\
 & DISTS $\downarrow$ & 0.2813 & 0.2615 & 0.2892 & 0.2455 & 0.2496 & 0.2483 & 0.2318 & 0.2625 & \bd{0.2169} & \rf{0.2079} \\
 & CLIP-I $\uparrow$ & 0.7674 & 0.8123 & 0.7526 & 0.7898 & 0.8117 & 0.8421 & 0.8537 & 0.8047 & \bd{0.8650} & \rf{0.8912} \\
 & NIQE $\downarrow$ & 4.1821 & \rf{4.0219} & \bd{4.0870} & 4.0970 & 5.6242 & 4.0666 & 4.2028 & 4.5963 & 4.3167 & 4.4363 \\
 & MANIQA $\uparrow$ & 0.3826 & 0.4338 & 0.3453 & 0.3333 & 0.3864 & 0.4400 & 0.4774 & 0.4269 & \bd{0.5557} & \rf{0.5811} \\
 & MUSIQ $\uparrow$ & 63.40 & 66.10 & 60.16 & 62.52 & 61.60 & 67.02 & 67.74 & 64.07 & \bd{71.90} & \rf{71.94} \\
\multirow{-9}{*}{LSDIR-Val} & CLIPIQA $\uparrow$& 0.5292 & 0.5965 & 0.5566 & 0.5848 & 0.5887 & 0.6708 & 0.6665 & 0.5692 & \bd{0.7114} & \rf{0.7274} \\ 
\bottomrule
\end{tabular}}
\end{table*}

%% file: tables/table2.tex
\begin{table*}[!ht]
\caption{Quantitative comparison with different state-of-the-art methods on real-world datasets. The best and second best results of each metric are highlighted in \rf{red} and \bd{blue}. LDM is not tested on RealLR200 dataset due to the limitations of the released version and GPU.}
\label{tab:real}
\vspace{-3mm}
\resizebox{1.0\textwidth}{!}{
\begin{tabular}{c|c|cccc|ccccccc}
\toprule
 &  & \multicolumn{4}{c|}{GAN-based Methods} & \multicolumn{6}{c}{Diffusion-based Methods} \\
\multirow{-2}{*}{Datasets} & \multirow{-2}{*}{Metrics} & BSRGAN\cite{zhang2021bsrgan} & Real-ESRGAN\cite{wang2021realesrgan} & DASR\cite{liang2022dasr} & FeMaSR\cite{chen2022femasr} & LDM\cite{rombach2022high} & StableSR\cite{wang2023exploiting} & DiffBIR\cite{lin2023diffbir} & PASD\cite{yang2023pixel} & SeeSR\cite{wu2023seesr} & SUPIR\cite{yu2024scaling}& Ours \\ \midrule
 & PSNR $\uparrow$ & \bd{24.75} & 24.15 & \rf{25.40} & 23.51 & 23.83 & 23.95 & 23.29 & 24.70 & 23.66 & 23.08 & 23.76 \\
 & SSIM $\uparrow$ & \bd{0.7401} & 0.7363 & \rf{0.7458} & 0.7088 & 0.6857 & 0.7240 & 0.6341 & 0.7213 & 0.6952 & 0.6470 & 0.7088 \\
 & LPIPS $\downarrow$ & \rf{0.2656} & \bd{0.2710} & 0.3134 & 0.2937 & 0.3159 & 0.2604 & 0.3567 & 0.2846 & 0.3004 & 0.3585 & 0.2865 \\
 & FID $\downarrow$ & 141.26 & 135.18 & 132.62 & 140.02 & 132.67 & 132.10 & 124.81 & \rf{120.97} & 125.07 & 128.45 & \bd{123.21} \\
 & DISTS  $\downarrow$ & 0.2124 & 0.2060 & 0.2202 & 0.2286 & 0.2215 & \rf{0.1990} & 0.2298 & \bd{0.2040} & 0.2218 & 0.2414 & 0.2136 \\
 & CLIP-I $\uparrow$ & 0.8635 & 0.8756 & 0.8768 & 0.8759 & 0.8681 & 0.8838 & 0.8770 & \bd{0.8895} & 0.8847 & 0.8703 & \rf{0.8956} \\
 & NIQE $\downarrow$ & 5.6362 & 5.8031 & 6.5454 & 5.7685 & 6.5139 & 6.6285 & 5.6486 & \rf{5.3805} & \bd{5.3968} & 6.77 & 5.5711 \\
 & MANIQA $\uparrow$ & 0.3764 & 0.3733 & 0.2461 & 0.3632 & 0.3635 & 0.3759 & 0.4602 & 0.4413 & \bd{0.5369} & 0.5041 & \rf{0.5544} \\
 & MUSIQ $\uparrow$ & 63.29 & 60.37 & 41.20 & 59.06 & 58.91 & 61.81 & 65.29 & 64.48 & \bd{69.70} & 62.21 & \rf{69.81} \\
\multirow{-9}{*}{RealSR} & CLIPIQA $\uparrow$ & 0.5116 & 0.4491 & 0.3200 & 0.5410 & 0.5680 & 0.5422 & 0.6558 & 0.5636 & \bd{0.6680} & 6.6573 & \rf{0.6765} \\ \midrule
 & PSNR $\uparrow$ & 26.39 & 26.28 & \rf{27.24} & 24.85 & 25.57 & \bd{26.70} & 24.74 & 26.29 & 25.90 & 24.60 & 25.31 \\
 & SSIM $\uparrow$ & 0.7739 & 0.7767 & \rf{0.7995} & 0.7247 & 0.7493 & \bd{0.7774} & 0.6155 & 0.7475 & 0.7434 & 0.6487 & 0.7351 \\
 & LPIPS $\downarrow$ & 0.2858 & \bd{0.2819} & 0.3099 & 0.3157 & 0.3379 & \rf{0.2698} & 0.4348 & 0.3320 & 0.3141 & 0.4137 & 0.3265 \\
 & FID $\downarrow$ & 155.60 & \bd{147.64} & 155.37 & 157.80 & 155.90 & 151.22 & 164.93 & 154.56 & \rf{146.98} & 152.19 & 152.90 \\
 & DISTS $\downarrow$ & 0.2144 & \bd{0.2089} & 0.2275 & 0.2239 & 0.2256 & \rf{0.2066} & 0.2691 & 0.2255 & 0.2298 & 0.2644 & 0.2347 \\
 & CLIP-I $\uparrow$ & 0.8807 & \bd{0.8935} & 0.8924 & 0.8912 & 0.8787 & 0.8921 & 0.8843 & 0.8837 & 0.8851 & 0.8854 & \rf{0.8960} \\
 & NIQE $\downarrow$ & 6.5327 & 6.6932 & 7.5868 & \rf{5.9035} & 7.1200 & 7.5441 & \bd{6.0035} & 6.7395 & 6.4766 & 7.69 & 6.6904 \\
 & MANIQA $\uparrow$ & 0.3425 & 0.3436 & 0.2845 & 0.3163 & 0.3451 & 0.3188 & 0.4553 & 0.4058 & \bd{0.5057} & 0.4663 & \rf{0.5255} \\
\multirow{-9}{*}{DrealSR} & MUSIQ $\uparrow$ & 57.17 & 54.27 & 42.41 & 53.71 & 53.73 & 51.36 & 61.46 & 56.12 & \bd{64.73} & 57.55 & \rf{66.27} \\
 & CLIPIQA $\uparrow$ & 0.5097 & 0.4520 & 0.3815 & 0.5642 & 0.5756 & 0.4905 & 0.6574 & 0.5682 & \bd{0.6895} & 0.6469 & \rf{0.6955} \\ \midrule
 & NIQE $\downarrow$ & 4.3656 & 4.1767 & 4.3180 & 4.6286 & - & 4.4676 & \rf{ 3.7673} & \bd{3.8752} & 4.2464 & 4.3250 & 4.4406 \\
 & MANIQA $\uparrow$ & 0.3671 & 0.3633 & 0.2937 & 0.4083 & - & 0.3622 & 0.4734 & 0.4314 & \bd{0.4845} & 0.4246 & \rf{0.5164} \\
 & MUSIQ $\uparrow$& 64.87 & 62.96 & 55.71 & 64.24 & - & 61.36 & 67.27 & 66.96 & \rf{68.37} & 63.13 & \bd{67.72} \\
\multirow{-4}{*}{RealLR200} & CLIPIQA $\uparrow$ & 0.5699 & 0.5409 & 0.4689 & 0.6548 & - & 0.5545 & 0.7022 & 0.6284 & \bd{0.6550} & 0.6114 & \rf{0.6930} \\ \bottomrule
\end{tabular}}
\end{table*}

%% file: tables/table3.tex
\begin{table*}[!htp]
\centering
\caption{Ablation studies of different semantic prompts on DIV2K-Val and RealSR datasets.}
\label{tab:prompt}
\vspace{-3mm}
\resizebox{1.0\textwidth}{!}{
\begin{tabular}{c|cccccc|cccccc}
\toprule
 & \multicolumn{6}{c|}{DIV2K-Val} & \multicolumn{6}{c}{RealSR} \\
\multirow{-2}{*}{Experiments} & PSNR $\uparrow$ & FID $\downarrow$ & LPIPS $\downarrow$ & CLIP-I $\uparrow$ & MUSIQ $\uparrow$ & CLIPIQA $\uparrow$ & PSNR $\uparrow$ & FID $\downarrow$ & LPIPS $\downarrow$ & CLIP-I $\uparrow$ & MUSIQ $\uparrow$ & CLIPIQA $\uparrow$ \\ \midrule
(1) & 19.56 & 41.40 & 0.3806 & 0.8814 & 66.20 & 0.6606 & 23.66 & 136.77 & 0.2936 & 0.8854 & 68.82 & 0.6637 \\
(2) & 19.38 & 38.75 & 0.3864 & 0.8898 & 66.11 & 0.6582 & 23.57 & 131.62 & 0.2905 & 0.8881 & 68.10 & 0.6451 \\
(3) & 19.09 & 39.28 & 0.4001 & 0.8915 & 65.77 & 0.6646 & 23.31 & 134.41 & 0.2970 & 0.8880 & 69.53 & 0.6591 \\
(4) & 19.28 & 39.32 & 0.3903 & 0.8955 & 66.35 & 0.6723 & 23.69 & 126.54 & 0.2827 & 0.8939 & 66.87 & 0.6224 \\
(5) & 19.50 & 45.63 & 0.3973 & 0.8663 & 64.90 & 0.6477 & 23.61 & 143.57 & 0.2993 & 0.8817 & 69.53 & 0.6723 \\
Ours & 19.18 & 38.07 & 0.3840 & 0.8982 & 69.27 & 0.7088 & 23.76 & 123.21 & 0.2865 & 0.8956 & 69.81 & 0.6765 \\ \bottomrule
\end{tabular}}
\end{table*}

%% file: tables/table4.tex
\begin{table*}[!htbp]
\centering
\caption{Ablation studies of loss functions on DIV2K-Val and RealSR datasets.}
\label{tab:loss}
\vspace{-3mm}
\resizebox{1.0\textwidth}{!}{
\begin{tabular}{cc|cccccc|cccccc}
\toprule
\multirow{2}{*}{$\mathcal{L}_{RGB}$} & \multirow{2}{*}{$\mathcal{L}_{FFT}$} & \multicolumn{6}{c|}{DIV2K-Val} & \multicolumn{6}{c}{RealSR} \\ 
 &  & PSNR $\uparrow$ & FID $\downarrow$ & LPIPS $\downarrow$ & CLIP-I $\uparrow$ & MUSIQ $\uparrow$ & CLIPIQA $\uparrow$ & PSNR $\uparrow$ & FID $\downarrow$ & LPIPS $\downarrow$ & CLIP-I $\uparrow$ & MUSIQ $\uparrow$ & CLIPIQA $\uparrow$ \\ \midrule
\XSolidBrush & \XSolidBrush & 19.19 & 38.80 & 0.3954 & 0.8946 & 67.03 & 0.6765 & 23.07 & 133.60 & 0.3025 & 0.8851 & 68.81 & 0.6564 \\
\Checkmark & \XSolidBrush & 19.16 & 39.24 & 0.3846 & 0.8964 & 68.63 & 0.7087 & 23.70 & 125.80 & 0.2995 & 0.8902 & 69.39 & 0.6644 \\
\XSolidBrush & \Checkmark & 19.15 & 39.14 & 0.3893 & 0.8935 & 68.19 & 0.7060 & 23.03 & 123.34 & 0.3028 & 0.8882 & 70.21 & 0.6817 \\
\Checkmark & \Checkmark & 19.18 & 38.07 & 0.3840 & 0.8982 & 69.27 & 0.7091 & 23.76 & 123.21 & 0.2865 & 0.8956 & 69.81 & 0.6765 \\ 
\bottomrule
\end{tabular}}
\end{table*}

%% file: tables/table5.tex
\begin{table*}[!htbp]
\centering
\caption{Ablation studies of attention module on DIV2K-Val and RealSR datasets.}
\label{tab:attention}
\vspace{-3mm}
\resizebox{1.0\textwidth}{!}{
\begin{tabular}{cc|cccccc|cccccc}
\toprule
Image & Strcuture & \multicolumn{6}{c|}{DIV2K-Val} & \multicolumn{6}{c}{RealSR} \\
Attention & Attention & PSNR $\uparrow$ & FID $\downarrow$ & LPIPS $\downarrow$ & CLIP-I $\uparrow$ & MUSIQ $\uparrow$ & CLIPIQA $\uparrow$ & PSNR $\uparrow$ & FID $\downarrow$ & LPIPS $\downarrow$ & CLIP-I $\uparrow$ & MUSIQ $\uparrow$ & CLIPIQA $\uparrow$ \\ \midrule
\XSolidBrush & \XSolidBrush & 18.87 & 44.41 & 0.4223 & 0.8803 & 68.22 & 0.7085 & 22.57 & 139.41 & 0.3255 & 0.8736 & 70.42 & 0.6740 \\
\Checkmark & \XSolidBrush & 19.29 & 37.80 & 0.3897 & 0.8868 & 65.15 & 0.6539 & 23.62 & 124.98 & 0.2787 & 0.8918 & 67.26 & 0.6093 \\
\XSolidBrush & \Checkmark & 19.63 & 45.45 & 0.3909  & 0.8678 & 65.73 & 0.6416 & 23.81 & 137.25 & 0.2898 & 0.8789 & 68.47 & 0.6393 \\
\Checkmark & \Checkmark & 19.18 & 38.07 & 0.3840 & 0.8982 & 69.27 & 0.7091 & 23.76 & 123.21 & 0.2865 & 0.8956 & 69.81 & 0.6765 \\ 
\bottomrule
\end{tabular}}
\end{table*}

%% file: tables/table_complexity.tex
\begin{table*}[!ht]
\centering
\caption{Analysis of the model's complexity. All methods are tested with an input image of size 512 × 512, and the inference time is measured on an A100 GPU.}
\label{tab:complexity}
\vspace{-3mm}
\resizebox{0.8\linewidth}{!}{
\begin{tabular}{c|ccccccc}
\toprule
Methods & Real-ESRGAN~\cite{wang2021realesrgan} & StableSR~\cite{wang2023exploiting} & DiffBIR~\cite{lin2023diffbir} & PASD~\cite{yang2023pixel} & SeeSR~\cite{wu2023seesr} & SUPIR~\cite{yu2024scaling} & Ours \\ \midrule
One-stage Training & \Checkmark & \XSolidBrush & \XSolidBrush & \XSolidBrush & \XSolidBrush & \XSolidBrush & \Checkmark \\
Trainable Params (M) & 16.70 & 153.27 & 378.94 & 609.52 & 751.66 & - & 997.80 \\ 
Inference Time (s) & 0.08 & 14.50 & 4.10 & 3.38 & 4.30 & 17.68 & 8.82 \\
\bottomrule
\end{tabular}}
\end{table*}